%% file: iclr2026_conference.tex
\documentclass{article} 

\ifdefined\pdfobjcompresslevel
\fi
\ifdefined\pdfvariable
  \pdfvariable objcompresslevel=0
\fi

\usepackage{iclr2026_conference}
\usepackage{newtxtext}
\input{math_commands.tex}

\usepackage{graphicx} 
\usepackage{hyperref}
\usepackage{url}
\usepackage{booktabs} 
\usepackage{dsfont}
\usepackage{amsmath}
\usepackage{placeins}  
\usepackage[ruled,vlined]{algorithm2e}

\title{CoTZero: Annotation-Free Human-Like \\Vision Reasoning  via Hierarchical Synthetic CoT}

\author{Chengyi Du \\
University of Electronic Science and Technology of China \\
Shanghai Artificial Intelligence Laboratory \\
\texttt{duchengyi1224@gmail.com} \\
\And
Yazhe Niu \thanks{corresponding author}\\
Shanghai Artificial Intelligence Laboratory \\
The Chinese University of Hong Kong MMLab \\
\texttt{niuyazhe314@outlook.com} \\
\And
Dazhong Shen \\
The College of Computer Science and Technology \\
Nanjing University of Aeronautics and Astronautics \\
\texttt{shendazhong@nuaa.edu.cn} \\
\AND
Luxin Xu \\
University of Electronic Science and Technology of China\\
\texttt{xul022332@gmail.com}
}

\iclrfinalcopy 
\begin{document}
\raggedbottom
\emergencystretch=2em
\maketitle
\begin{abstract}

Recent advances in vision–language models (VLMs) have markedly improved image–text alignment, yet they still fall short of human-like visual reasoning. A key limitation is that many VLMs rely on surface correlations rather than building logically coherent structured representations, which often leads to missed higher-level semantic structure and non-causal relational understanding, hindering compositional and verifiable reasoning. To address these limitations by introducing human models into the reasoning process, we propose CoTZero, an annotation-free paradigm with two components: \textbf{(i) a dual-stage data synthesis approach and (ii) a cognition-aligned training method}. In the first component we draw inspiration from neurocognitive accounts of \textit{compositional productivity} and \textit{global-to-local analysis}. In the bottom-up stage, CoTZero extracts atomic visual primitives and incrementally composes them into diverse, structured question–reasoning forms. In the top-down stage, it enforces hierarchical reasoning by using coarse global structure to guide the interpretation of local details and causal relations. In the cognition-aligned training component, built on the synthesized CoT data, we introduce \textbf{Cognitively Coherent Verifiable Rewards} (CCVR) in Reinforcement Fine-Tuning (RFT) to further strengthen VLMs' hierarchical reasoning and generalization, providing stepwise feedback on reasoning coherence and factual correctness. Experiments show that CoTZero achieves an F1 score of 83.33\% on our multi-level semantic inconsistency benchmark with lexical-perturbation negatives, across both in-domain and out-of-domain settings. Ablations confirm that each component contributes to more interpretable and human-aligned visual reasoning.
\end{abstract}
\input{section/intro}
\input{section/related_work}
\input{section/method.tex}

\input{section/experiment}
\input{section/conclusion}
\begingroup
\raggedright
\sloppy
\emergencystretch=3em
\bibliography{iclr2026_conference}
\bibliographystyle{iclr2026_conference}
\endgroup
\clearpage
\appendix
\input{section/Appendix}

\end{document}

%% file: math_commands.tex
\usepackage{amsmath,amsfonts,bm}

\def\eqref#1{equation~\ref{#1}}

\def\1{\bm{1}}

\DeclareMathAlphabet{\mathsfit}{\encodingdefault}{\sfdefault}{m}{sl}
\SetMathAlphabet{\mathsfit}{bold}{\encodingdefault}{\sfdefault}{bx}{n}



%% file: section/intro.tex
\section{Introduction}

\begin{figure}
    \centering
    \includegraphics[width=1.0\linewidth]{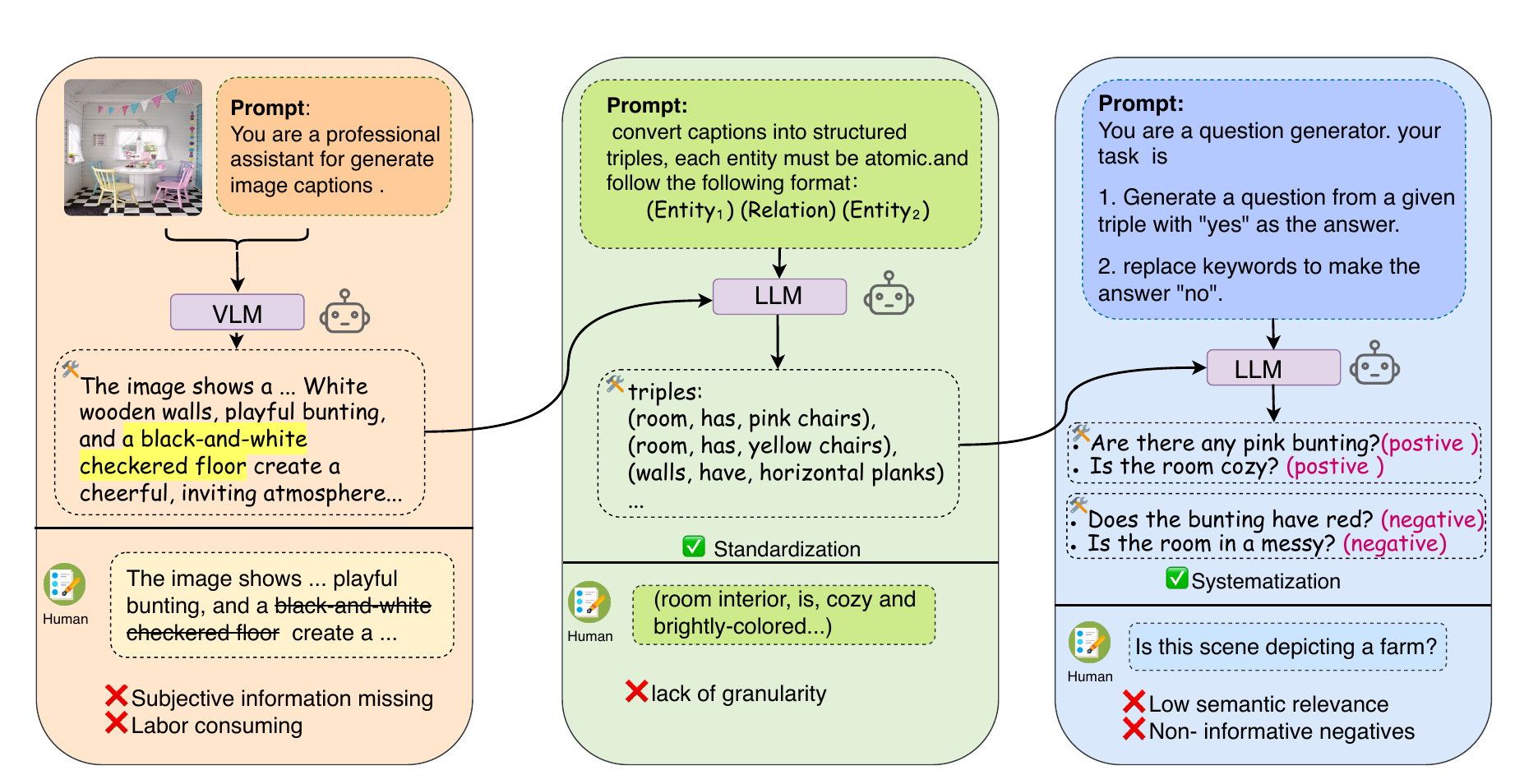}
    \caption{Our annotation-free data generation pipeline. Starting from external image inputs, a VLM produces rich captions, which are structured into (entity, relation, entity) triples by an LLM. These triples are then transformed into yes/no QA pairs through controlled prompting. This pipeline enables scalable, fine-grained, and semantically consistent supervision without human annotation.}
    \label{fig:datageneration}
    \vspace{-15pt} 
\end{figure}

\begin{figure}[t]
    \centering
    \begin{minipage}{0.50\linewidth}
        \centering
        \includegraphics[width=\linewidth,height=0.32\textheight,keepaspectratio]{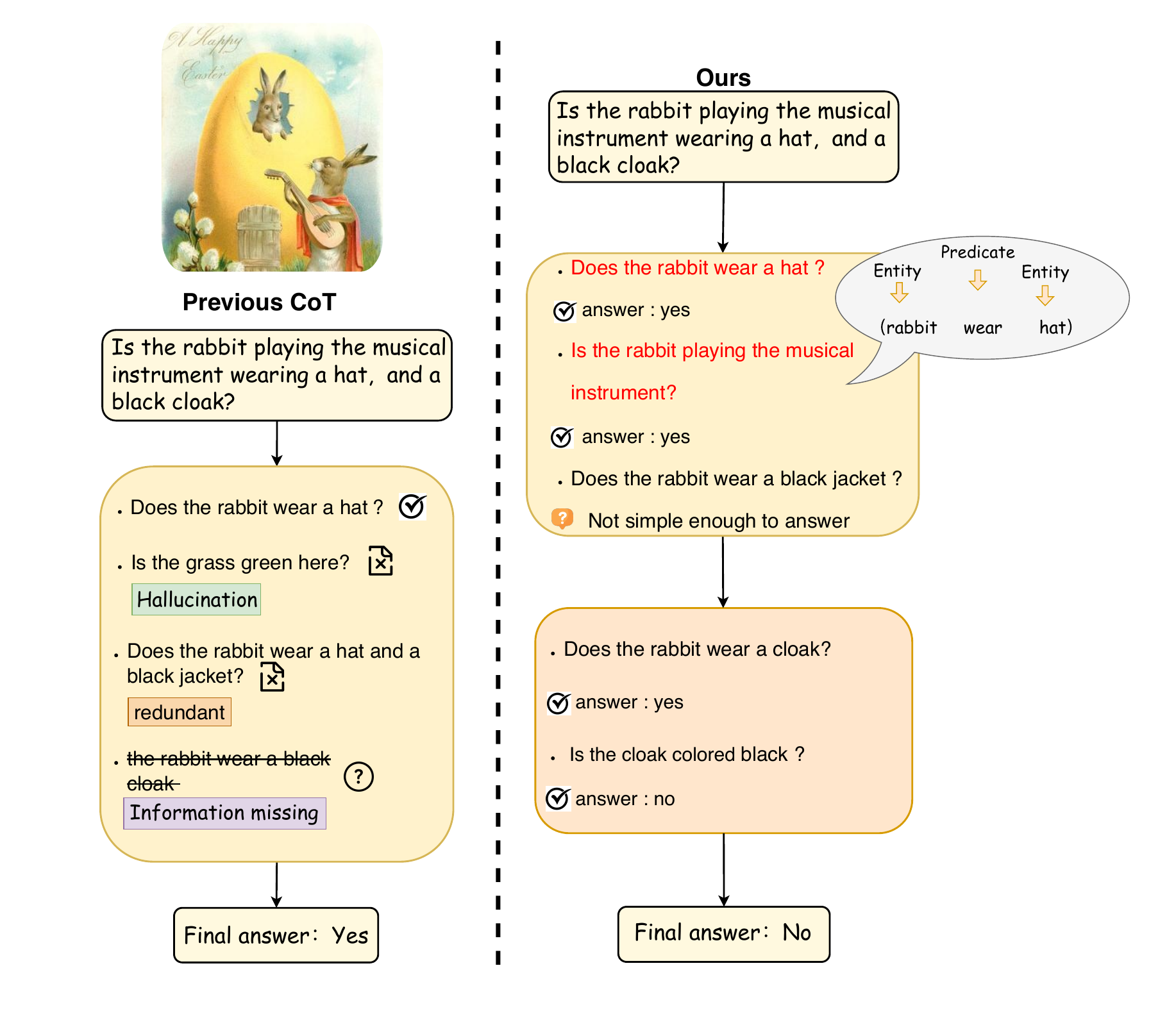}
        \caption{Comparison between our CoT process and previous CoT process.}
        \label{cot_com}
        \vspace{-8pt}
    \end{minipage}
    \hfill
    \begin{minipage}{0.36\linewidth}
        \raggedright In contrast, we design a rule-based reward function that jointly considers edit distance and semantic similarity between the model-generated reasoning chain and the reference CoT, thereby capturing both semantic coherence and hierarchical alignment. This reward design provides a learning signal that closely mirrors the way humans incrementally construct and
        validate their understanding of complex visual scenes. We detail this process in
        Section~\ref{sec:Reinforcement Learning with Cognitively Coherent Verifiable Rewards}.The model first undergoes a cold-start phase on a public dataset to instill a cognitive blueprint.
        \vspace{-20pt}
    \end{minipage}

\end{figure}

Vision-Language Models (VLMs) \cite{Visualinstructiontuning, OpenAI_GPT4V_SystemCard_2023, LLaVA2025, zhu2025internvl3exploringadvancedtraining, bai2025qwen3vltechnicalreport} have achieved notable progress in tasks such as image captioning \cite{dong2024benchmarkingimprovingimagecaption, wang2024tarsierrecipestrainingevaluating, lu2025benchmarkinglargevisionlanguagemodels}, visual question answering(VQA) \cite{10.1016/j.cosrev.2023.100548, Antol_2015_ICCV, mañas2024improvingautomaticvqaevaluation}, and text-to-image \cite{Narasimhaswamy_2024, li2024cosmicmantexttoimagefoundationmodel} generation. Yet they still struggle with the fundamental challenge of visual reasoning-particularly when faced with complex scenarios demanding hierarchical analysis \cite{huang2025visionlanguagemodelsstruggle}. While these models excel at surface-level image-text associations, their reasoning capabilities remain far behind human-like performance. This persistent gap arises because current VLMs rely on statistical pattern recognition rather than the construction of human-like, causally structured models of the world. 
To bridge the gap between these human traits and AI, some works attempt to teach models through synthetic data. However, as illustrated in Figure~\ref{fig:datageneration}, current visual reasoning data generation frequently involves subjective manual annotation or yields unstructured, low-granularity knowledge. These methods are not only labor-intensive but also fail to capture the deep causal structure required for human-like reasoning. Other efforts focus on direct model generation; yet, prevalent Chain-of-Thought (CoT) methods for visual tasks primarily generate linear or unstructured thought sequences. As illustrated in Figure~\ref{cot_com}, this lack of hierarchy causes reasoning to suffer from hallucination, redundancy, and information missing. Without a structured mechanism to mirror the Bayesian brain’s drive to minimize surprise (prediction error) through stagewise verification, these "unstructured" thoughts remain unreliable and lack the interpretability necessary for complex visual scenarios.

Human intelligence is distinguished by two profound cognitive capabilities. 
First, In human cognition, productivity is at the core of compositionality \cite{lake2016buildingmachineslearnthink, hockett1960origin}, enabling the mind to construct an infinite number of thoughts and diverse reasoning paths from a finite set of primitives. This flexibility enables humans to navigate multiple itinerant "thought trajectories" that, while varied, are all anchored by causal coherence to reach the same correct conclusion. Second, human perception follows a global-to-local trajectory \cite{Hybridimages, NAVON1977353}, where an initial analysis of global layouts guides the interpretation of fine-grained local details. Motivated by these two cognitive principles, we introduce CoTZero, an annotation-free paradigm that consists of a dual-stage data synthesis approach and a cognition-aligned training method. Drawing on the "analysis-by-synthesis" paradigm \cite{Analysis-by-Synthesis, lake2016buildingmachineslearnthink, 10.1145/1464052.1464081}, our data approach represents visual scenes by modeling the causal process that generated question-reasoning pairs. Specifically, in the bottom-up stage, CoTZero operationalizes compositional productivity \cite{lake2016buildingmachineslearnthink, hockett1960origin, Marc2002-HAUTFO} — the human capacity to construct infinite representations from a finite set of primitives. By extracting atomic visual elements and structured triples, our data approach incrementally composes these units into diverse question forms, enabling the emergence of potentially unbounded reasoning traces. In the top-down stage, CoTZero adheres to the cognitive principle of global-to-local analysis \cite{Hybridimages, NAVON1977353}, a trajectory supported by the hierarchical architecture of human vision. Neuroscientific studies \cite{Felleman1991DistributedHierarchy, BULL202117} reveal that the human visual cortex processes information through successive stages, progressing from low-level feature detection in the primary visual cortex (V1) to high-level integration and category recognition in the inferotemporal cortex (IT), --namely “seeing the forest before the trees” \cite{NAVON1977353}. By operationalizing this hierarchical logic, CoTZero systematically decomposes complex queries into nested, verifiable subproblems, ensuring that the global structure of a scene provides the necessary causal constraints to guide the interpretation of atomic components. Integrating this dual-stage data synthesis approach, CoTZero ensures global coherence and precise local understanding, thereby overcoming the unstructured and pattern-driven processing limitations of current VLMs. 

Building on the synthesized hierarchical CoT trajectories produced by our dual-stage approach, we propose a cognition-aligned training method that enhances reinforcement fine-tuning with \textbf{Cognitively Coherent Verifiable Rewards} (CCVR). Unlike conventional approaches that evaluate only final outputs like using IoU scores in object grounding \cite{Du2025MultiObjectGrounding}, accuracy in visual question answering \cite{mañas2024improvingautomaticvqaevaluation}, or success rates in navigation tasks \cite{zhao2023mindgapimprovingsuccess}, recent work has begun to adopt process-supervised reward models (PRMs) that provide intermediate-step supervision for logical reasoning \cite{shao2024deepseekmathpushinglimitsmathematical}, these PRMs typically rely on LLM-based judges or learned reward models for scoring

 It is then trained with Supervised Fine-Tuning (SFT) on the dataset generated by our dual-stage approach. 
Subsequently, the model is reinforcement fine-tuned using the Group Relative Policy Optimization (GRPO) algorithm \cite{shao2024deepseekmathpushinglimitsmathematical}, with Cognitively Coherent Verifiable Rewards (CCVR) as the reward mechanism. This CCVR-guided GRPO stage improves accuracy by 15.90\% over the baseline. We hope these insights and advancements unlock deeper compositional reasoning in VLMs, bridging the gap between powerful data-driven training and hierarchical human-like reasoning.

%% file: section/related_work.tex
\section{Related Work}
\subsection{Advancing VLM Problem-Solving Performance}
In the evolution of VLMs, Chain-of-Thought (CoT) technology is regarded as the core driving force for achieving the transition from foundational "visual perception" to high-order "cognitive reasoning" \cite{zhou2025perceptioncognitionsurveyvisionlanguage}. While early research primarily relied on prompt engineering \cite{zhang2024multimodalchainofthoughtreasoninglanguage, wei2023chainofthoughtpromptingelicitsreasoning} to stimulate reasoning potential, the current research focus has shifted toward enhancing problem-solving abilities via training. Specifically, training methodologies have advanced from initial imitation learning, such as the text-visual dependency constraints introduced by Multimodal-CoT \cite{zhang2024multimodalchainofthoughtreasoninglanguage}, to phased curriculum learning exemplified by the decomposition-alignment-integration stages of LLaVA-CoT \cite{xu2025llavacotletvisionlanguage} and LlamaV-o1 \cite{thawakar2025llamavo1rethinkingstepbystepvisual}. Recent efforts have further adopted preference learning algorithms like GRPO \cite{shao2024deepseekmathpushinglimitsmathematical} and DPO \cite{rafailov2024directpreferenceoptimizationlanguage} to optimize logical consistency and align reasoning paths with visual facts. Furthermore, reasoning mechanisms are transitioning from traditional linear greedy search toward non-linear structures based on Tree of Thoughts (ToT) \cite{10.5555/3666122.3666639} and Monte Carlo Tree Search (MCTS), which allow models to perform backtracking and multi-path exploration. The latest progress emphasizes a dynamic reasoning loop ("Think with Image") \cite{su2025thinkingimagesmultimodalreasoning,zhou2025perceptioncognitionsurveyvisionlanguage}, where the model continuously re-examines visual evidence according to evolving inferential needs through endogenous attention refocusing \cite{gao2025interleavedmodalchainofthought, yang2025lookbackimplicitvisualrefocusing, qi2025cogcomvisuallanguagemodel} or exogenous tool calls \cite{gupta2022visualprogrammingcompositionalvisual}. thereby effectively alleviating hallucinations in long-chain reasoning.

\subsection{Reinforcement Learning and Reward Mechanism Design for Reasoning in VLMs}
While pre-trained models perform well on many tasks, they often fall short in complex reasoning and human-aligned generation; thus, post-training has progressed from supervised fine-tuning (SFT) to preference-based \cite{ouyang2022traininglanguagemodelsfollow,xu2025largereasoningmodelssurvey}, where Reinforcement Learning from Human Feedback (RLHF) optimizes policies via a learned reward model (e.g., PPO \cite{schulman2017proximalpolicyoptimizationalgorithms}), and DPO \cite{rafailov2024directpreferenceoptimizationlanguage} provides a simpler alternative by directly learning from pairwise preferences without explicit reward modeling. Central to this evolution is the design of the reward model, which has shifted from providing sparse, end-to-end feedback to delivering detailed, process-oriented supervision. Early alignment efforts relied on Outcome Reward Models (ORM) \cite{luong2024reftreasoningreinforcedfinetuning, kazemnejad2025vinepporefiningcreditassignment}, where feedback is concentrated solely on the final solution; however, this approach faces a severe credit assignment problem \cite{xu2025largereasoningmodelssurvey}, as the model struggles to identify which specific intermediate steps contributed to the final result. To overcome this bottleneck, the research frontier has moved toward Process Reward Models (PRM) \cite{hwang2024selfexploreenhancingmathematicalreasoning, wang2024mathshepherdverifyreinforcellms}, which provide dense, step-wise rewards that encourage models to master human-like reasoning trajectories through trial-and-error. 

%% file: section/method.tex
\section{Method}
\subsection{Overview}

\subsection{dual-stage data synthesis approach}

\label{sec:Annotation-free Hierarchical CoT Generation}
As shown in Figure~\ref{fig:datageneration}, dual-stage data synthesis approach requires only image inputs, thereby eliminating the need for additional human annotations. This automated process efficiently produces structured question-reasoning data tailored for enhancing the vision reasoning capabilities of models. We jointly leverage visual and language models to generate structured chain-of-thought (CoT) training data. This helps the model better capture fine-grained reasoning steps that are often overlooked in long-form multimodal inputs. This enhances the model’s capability in tasks such as visual question answering (VQA) and multimodal reasoning.

At the initial stage of our pipeline, a Vision-Language Model (VLM) is employed to extract semantically rich captions from raw visual inputs. These captions are then processed by a LLM to generate triples, where each triple is structured as

\begin{equation}
(\mathcal{E}_i, \mathcal{R}_{\text{attr}}/\mathcal{R}_{\text{verb}}/\mathcal{R}_{\text{locate}}/\mathcal{R}_{\text{exist}}, \dots, \mathcal{E}_j)
\end{equation}

Each triple consists of two entities $\mathcal{E}_{i,j}$ and a relation $\mathcal{R}$ that links them, representing the key attributes and relationships within the visual input. Our triples typically contain the simplest relationships possible, maintaining a structured format. This simplicity ensures that each triple remains easily interpretable and directly relevant to downstream tasks. By avoiding overly complex or nested relationships, our approach facilitates straightforward reasoning and efficient generation process. After obtaining triples, we first generate the simplest atomic questions based on individual triples, capturing basic attributes or object relations. These atomic questions serve as the foundational building blocks. To construct semantically relevant negative samples, we further generate lexically altered negatives by replacing key lexical elements—typically the entity or attribute that determines the answer—with alternatives that render the question incompatible with the image content. This process ensures that the negative questions remain grammatically correct and contextually plausible, yet contradict the visual semantics, thereby enhancing the model’s ability to perform fine-grained visual reasoning.See Appendix~\ref{app:Data_Generation} for details.

\begin{figure}
    \centering
    \includegraphics[width=1.0\linewidth]{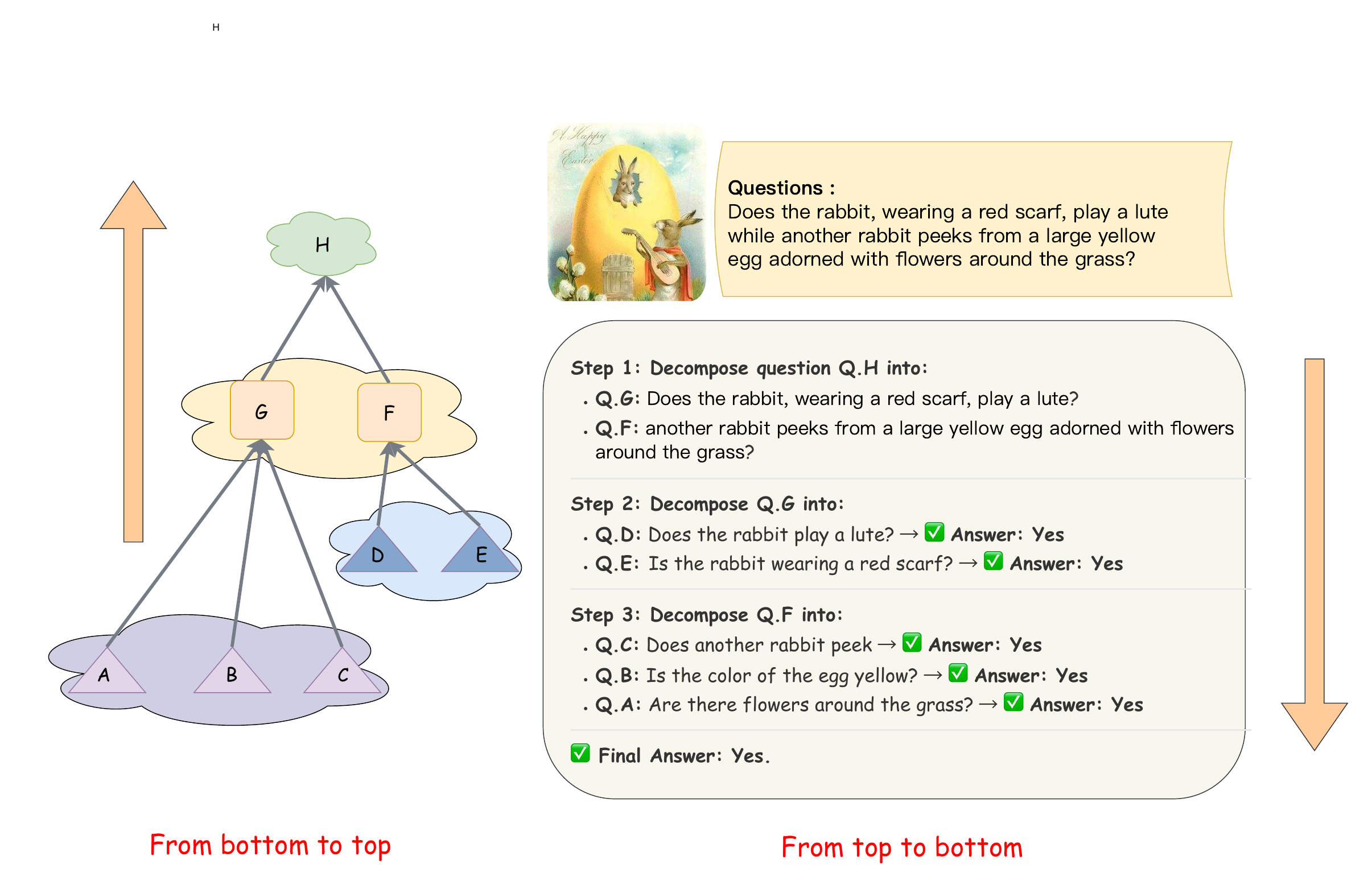}
    \caption{Atomic-level questions derived from image captions are merged bottom-up into higher-level reasoning chains based on semantic similarity, forming a question hierarchy. In turn, complex questions are decomposed top-down to generate training data with multi-granularity supervision.}
    \label{fig:tree}
\end{figure}

\textbf{Bottom-Up Merging to Build the Question Tree.} 
We start by generating atomic questions from basic attributes and relationships extracted from data. These atomic questions serve as the fundamental units. Using a similarity-based merging strategy, we iteratively group similar questions to form intermediate-level questions. This merging process continues until we obtain complex, long-text questions representing high-level reasoning tasks.The merging criterion is determined by the embedding similarity between questions, calculated using pre-trained language models. Specifically, we compute the cosine similarity between sentence embeddings using the following formula:
\begin{equation}
\text{cosine}(s_i, s_j) = \frac{s_i \cdot s_j}{\|s_i\| \|s_j\| + \epsilon}
\end{equation}

Here, $s_i$ and $s_j$ represent the embedding vectors of two questions, and the small constant $\epsilon = e^{-9}$ is added to prevent division by zero. To ensure diversity, the number of questions merged at each step is chosen randomly from a predefined range. This progressive merging forms a hierarchical question tree, where each level encapsulates a different granularity of information.

\textbf{Top-Down Decomposition to Generate Training Data.}
Once the question tree is constructed, we utilize it in a reverse manner to generate data, as illustrated in Figure~\ref{fig:tree}. Starting from the complex long-text questions at the top, we systematically decompose them into intermediate and atomic questions. This top-down decomposition mirrors the reasoning process where complex queries are broken down into simpler subproblems. By representing the reasoning path in a hierarchical structure, the model learns to navigate from abstract, high-level concepts to concrete, fundamental details, thereby enhancing its capacity for comprehensive question understanding.

By leveraging this dual approach — constructing questions in a bottom-up manner while generating training data in a top-down fashion — we ensure that the model is exposed to both complex reasoning and fundamental atomic units. This combination fosters robust question-reasoning capabilities, allowing the model to better handle multi-step reasoning and long-text analysis.

\subsection{a cognition-aligned training method}
\label{sec:Reinforcement Learning with Cognitively Coherent Verifiable Rewards}
While preliminary SFT provides basic reasoning, it often fails on complex queries by relying on superficial patterns. We thus adopt RFT with CCVR, leveraging our hierarchical CoT trajectories to ensure structured and logically consistent reasoning.

\subsubsection{GRPO}
We utilize the Group Relative Policy Optimization (GRPO) \cite{shao2024deepseekmathpushinglimitsmathematical} method to fine-tune large language models (LLMs) for enhanced reasoning performance in complex tasks. 
GRPO is designed to optimize model behavior by leveraging comparisons among multiple responses generated from the same prompt, rather than solely evaluating individual outputs.

The GRPO training begins with the generation of multiple response sequences for each input prompt.
These responses are then evaluated using a reward function, which assigns scores based on the quality and relevance of each response.
To capture relative performance, the advantage of each response is computed by normalizing the reward values within the group, following the formula:

\begin{equation}
\hat{A}^{i,t} = \frac{r_i - \mu_r}{\sigma_r}
\end{equation}

where $r_i$ denotes the reward of the $i$-th response, while $\mu_r$ and $\sigma_r$ represent the mean and standard deviation of the rewards across the group, respectively. This normalized advantage helps to highlight how each response compares to others in the same batch. To prevent the model from deviating excessively from the reference policy, the KL divergence between the current policy and the reference policy is estimated using the following formula:

\begin{equation}
D_{\mathrm{KL}}[\pi_\theta \| \pi_{\text{ref}}] = \frac{\pi_{\text{ref}}(o_{i,t} \mid q, o_{i,<t})}{\pi_\theta(o_{i,t} \mid q, o_{i,<t})} - \log \frac{\pi_{\text{ref}}(o_{i,t} \mid q, o_{i,<t})}{\pi_\theta(o_{i,t} \mid q, o_{i,<t})} - 1,
\end{equation}

The overall loss function in GRPO combines the advantage function and the KL divergence penalty as follows:

{\scriptsize
\begin{equation}
\mathcal{L}_{\text{GRPO}}(\theta) = - \mathbb{E}_{(q, \{o_i\})} \left[ \sum_{i=1}^{G} \sum_{t=1}^{|o_i|} \left( \frac{\pi_\theta(o_{i,t} \mid q, o_{i,<t})}{\pi_{\text{ref}}(o_{i,t} \mid q, o_{i,<t})} \hat{A}_{i,t} - \beta D_{\text{KL}}[\pi_\theta \| \pi_{\text{ref}}] \right) \right]
\end{equation}
}

Here, $\beta$ acts as a hyperparameter controlling the penalty for divergence, ensuring that the updated policy remains close to the reference policy. This approach facilitates more stable optimization, allowing the model to adapt to complex reasoning tasks without sacrificing consistency or performance. By integrating the relative comparison of multiple outputs and maintaining a balanced update through KL regularization, GRPO enhances the robustness and efficacy of LLM fine-tuning.

\subsubsection{Cognitively Coherent Verifiable Rewards (CCVR)}

Building on the GRPO framework, we introduce  Cognitively Coherent Verifiable Rewards (CCVR), designed to evaluate both the final output and intermediate reasoning steps. The complete optimization procedure is summarized in Algorithm~\ref{alg:ccvr}. Unlike GRPO, which typically focuses on optimizing generative models through reinforcement learning by maximizing a reward function that evaluates the quality of final responses. Our approach incorporates structured reasoning evaluation directly into the reward function. Our rule-based reward function decomposes the entire reasoning process into structured steps, using predefined rules to capture both the accuracy of intermediate reasoning and the logical consistency of the final output. 
To comprehensively assess response quality, the model integrates three core components: Format Reward, Answer Reward, and Process Reward. These components address structural correctness, output accuracy, and reasoning coherence, respectively, ensuring a balanced evaluation that prioritizes both accurate answers and consistent reasoning. The overall reward $r$ is calculated as a combination of the three components:

\vspace{-8pt}
\begin{equation}
r = \lambda_{\mathrm{format}}\cdot r_{\text{format}} + \lambda_{\mathrm{format}}\cdot r_{\text{answer}} + \lambda_{\mathrm{format}}\cdot r_{\text{process}}
\vspace{-4pt}
\end{equation}

Here, $r_{\text{format}}$, $r_{\text{answer}}$, and $r_{\text{process}}$ represent the format, answer, and process rewards, respectively. The $\lambda_{\mathrm{format}}$, $\lambda_{\mathrm{format}}$, and $\lambda_{\mathrm{format}}$ control the relative importance of each component. This combined reward structure ensures that the model not only generates accurate final answers but also adheres to structured reasoning and formatting.

\textbf{Format and Answer Rewards}. The Format Reward evaluates structural correctness by verifying the presence and sequence of \texttt{<think>} and \texttt{<answer>} tags, ensuring model interpretability. The Answer Reward ensures alignment with ground-truth labels by rewarding explicit, conclusive answers. To mitigate reward hacking, it prioritizes single, unambiguous conclusions over contradictory outputs, better aligning the model with human-like decision-making

\textbf{Process reward}. The process reward is a critical component that evaluates the coherence and logical consistency of the reasoning steps. In structured reasoning tasks, maintaining a consistent logical pathway from input to final output is essential. This reward component quantifies the alignment between the generated reasoning process and a reference reasoning chain. It is specifically designed to capture both semantic similarity and logical sequence coherence, which are crucial for accurately modeling multi-step reasoning tasks. The process reward combines two sub-scores with linear weighting: the Semantic Score and the Edit Distance Score:
\begin{equation}
r_{\text{process}} = \lambda \cdot S_{\text{semantic}} + (1 - \lambda) \cdot S_{\text{edit}}
\vspace{-5pt}
\end{equation}

Here, $\lambda$ controls the balance between the semantic similarity and the edit distance components. The semantic score quantifies the similarity between the generated reasoning steps and the reference reasoning steps. To calculate this score, we first encode the sentences from both the generated and reference reasoning using an embedding model, which converts each sentence into a representation. We then compute the similarity between each pair of sentences using cosine similarity. The semantic score is calculated as the proportion of sentence pairs whose similarity exceeds a threshold $\delta$:

\begin{equation}
S_{\text{semantic}} = \frac{\sum_{i=1}^{m} \sum_{j=1}^{n} \mathds{1}\left(\text{Similarity}(x_i, y_j) \geq \delta\right)}{\max(m, n)}
\end{equation}

In this formula, $m$ and $n$ denote the number of sentences in the generated reasoning and reference reasoning, respectively, and $\mathds{1}$ is an indicator function that returns 1 if the similarity exceeds $\delta$, and 0 otherwise. This score captures the extent to which the generated reasoning steps are semantically aligned with the expected logical flow, thereby ensuring that both content accuracy and reasoning coherence are assessed.
To compute the soft edit distance between a generated reasoning chain $G = \{g_1, \ldots, g_m\} $and a reference chain $T = \{t_1, \ldots, t_n\}$, we define a dynamic programming matrix  $D \in \mathbb{R}^{(m+1) \times (n+1)}$, where $D_{i,j}$ represents the minimum number of operations required to align the first $i$ sentences of $G$ with the first $j$ sentences of $T$.We initialize the borders as:
\begin{equation}
D_{0,j}=j,\ \forall j\in[0,n], \qquad
D_{i,0}=i,\ \forall i\in[0,m].
\end{equation}

The recurrence is defined for $i>0$ and $j>0$ (i.e., excluding the boundary cases).
\begin{equation}
\begin{aligned}
D_{i,j} &=
\begin{cases}
D_{i-1,j-1}, & \text{if } \mathrm{sim}(g_i, t_j) \ge \theta, \\
\min\!\left\{
\begin{array}{l}
D_{i-1,j} + 1 \\
D_{i,j-1} + 1 \\
D_{i-1,j-1} + 1
\end{array}
\right., & \text{otherwise}
\end{cases}
\qquad
S_{\text{edit}} &= 1 - \frac{D_{m,n}}{n}.
\end{aligned}
\end{equation}

Here, $\text{sim}(g_i, t_j) \in [0,1]$ denotes the cosine similarity between sentence embeddings, and $\theta$ is the similarity threshold for considering two sentences semantically equivalent. $D_{m,n}$ is the total soft edit distance, and $n$ is the number of reference sentences used for normalization.

\begin{algorithm}[t]
\label{alg:ccvr}
\caption{Reinforcement Fine-Tuning with Cognitively Coherent Verifiable Rewards}
\KwIn{
Initial policy model $\pi_\theta$ (from a pretrained VLM);\\
Dataset $\mathcal{D} = \{(x_i, y_i, r^*_i)\}$, where $x_i$: prompt, $y_i$: ground-truth answer, $r^*_i$: reference reasoning chain
}
\KwOut{Updated policy model $\pi_\theta$}

\For{each training step}{
    Sample $(x, y, r^*)$ from dataset $\mathcal{D}$\;
    Generate $K$ trajectories $\{\tau_k\}_{k=1}^K$ using $\pi_\theta(\tau \mid x)$\;

    \For{each trajectory $\tau_k$}{
        Compute format score $s_{\text{format}}$ based on presence and order of \texttt{<think>} and \texttt{<answer>}\;

        Compute answer score $s_{\text{answer}}$ by comparing answer in \texttt{<answer>} section with $y$\;

        Extract reasoning steps from \texttt{<think>} section and compute process score:\;
        - Semantic similarity $s_{\text{sem}}$ via Sentence-BERT\;
        - Edit similarity $s_{\text{edit}}$ via normalized edit distance\;
        - $s_{\text{proc}} \leftarrow \lambda s_{\text{sem}} + (1 - \lambda) s_{\text{edit}}$\;

        Compute final reward:\;
        $s_k \leftarrow \lambda_{\mathrm{format}} s_{\text{format}} + \lambda_{\mathrm{answer}} s_{\text{answer}} + \lambda_{\mathrm{proc}} s_{\text{proc}}$\;
    }

    Compute baseline reward: $\bar{s} \leftarrow \frac{1}{K} \sum_{k=1}^K s_k$\;

    Compute policy gradient:\;
    $\nabla_\theta \mathcal{L} \leftarrow \frac{1}{K} \sum_{k=1}^K (s_k - \bar{s}) \nabla_\theta \log \pi_\theta(\tau_k \mid x)$\;

    Update $\pi_\theta$ using gradient descent\;
}
\Return{$\pi_\theta$}
\end{algorithm}

\FloatBarrier

%% file: section/experiment.tex
\section{Experiment}
\subsection{Experiment Settings}
\textbf{Dataset and Metrics}. We first constructed a training set of 1,000 image-question pairs, which was subsequently expanded to 4,000 samples to achieve consistent performance improvements. For evaluation, we curated a multi-source benchmark consisting of 100 high-quality samples uniformly selected from five datasets. This test set is structured with multiple difficulty levels and fine-grained, lexically altered negative samples to assess the model’s reasoning robustness under challenging conditions. Performance is measured using Accuracy (ACC) and F1 Score across total, in-domain, and out-of-domain settings. Comprehensive details on the composition of the data set, source distributions and construction protocols are provided in Appendix~\ref{app:Dataset_Details}

\subsection{Main Results}
\begin{table}[t]
\begin{center}
\caption{Performance comparison under different training strategies.}
\label{main-results}
\begin{tabular}{lcccccc}
\toprule
\textbf{Method} & 
\multicolumn{2}{c}{\textbf{Total}} & 
\multicolumn{2}{c}{\textbf{In-domain}} & 
\multicolumn{2}{c}{\textbf{Out-of-domain}} \\
\cmidrule(lr){2-3} \cmidrule(lr){4-5} \cmidrule(lr){6-7}
& ACC & F1 & ACC & F1 & ACC & F1 \\
\midrule
Qwen-VL 2.5 (3B) & 65.31 & 71.98 & 56.41 & 66.67 & 67.52 & 73.30 \\
+ SFT (LoRA) & 72.04 & 75.47 & 62.16 & 66.67 & 74.50 & 77.65 \\
+ Cold Start & 66.67 & 73.04 & 51.35 & 59.09 & 70.47 & 76.34 \\
+ CCVR & 77.42 & 80.19 & 81.08 & 82.05 & 76.51 & 79.77 \\
+ Cold Start + SFT + CCVR (\textit{Best}) & \textbf{81.11} & \textbf{81.95} & 76.68 & 75.68 & 81.21 & 83.33 \\
\bottomrule
\end{tabular}
\end{center}
\end{table}

Table~\ref{main-results} compares different training strategies on our annotation-free, hierarchically structured CoTZero dataset (see Sec.~\ref{experiment_config} for the experimental setup and evaluation protocol). Conventional fine-tuning alone yields only moderate improvements and remains insufficient for robust and interpretable reasoning. In contrast, leveraging our training method with CCVR gains in overall performance and produces consistent improvements across total, in-domain, and out-of-domain evaluations, indicating better generalization and more reliable reasoning behavior.

Compared to the baseline, our full method improves total F1 by nearly 10\%, with over 20\% improvement in in-domain accuracy. Out-of-domain F1 also rises substantially, demonstrating enhanced generalization. These results underscore the effectiveness of our synthetic reasoning data in teaching structured, multi-step thought processes. Moreover, CCVR further reinforce the model’s ability to produce verifiable and logically coherent reasoning chains. Overall, our framework outperforms all other configurations, validating the synergy between structured data generation and reasoning-aware optimization. Additional qualitative results are provided in Appendix~\ref{app:Additional_Quantitative_Results}.

\subsection{Ablation Results}

\subsubsection{Ablation Study of the Synthesis Method}
\vspace{-8pt} 
\begin{table}[htbp]
\centering
\small
\caption{Impact of Different Negative Sample Strategies for CoT Data Generation.}
\label{ablation-neg}
\begin{tabular}{lcc}
\toprule
\textbf{Strategy Setting} & \textbf{ACC (\%)} & \textbf{F1 (\%)} \\
\midrule
Baseline (no Negative Samples) & 65.30 & 71.98 \\
Cross-image negatives & 51.61 & 67.68 \\
Lexically altered negatives & \textbf{65.59} & \textbf{74.80} \\
\bottomrule
\end{tabular}
\end{table}

\begin{table}[htbp]
\centering
\small
\caption{Effect of Atomic QA Composition Strategies}
\label{ablation-comp}
\begin{tabular}{lcc}
\toprule
\textbf{Atomic QA Composition Strategy} & \textbf{ACC (\%)} & \textbf{F1 (\%)} \\
\midrule
Baseline (no Composition) & 65.31 & 74.98 \\
Fixed template combination & 77.42 & 79.21 \\
Dual-stage data synthesis approach & \textbf{81.21} & \textbf{84.09} \\
\bottomrule
\end{tabular}

\end{table}

To understand the contribution of our proposed training data components, we conduct a comprehensive ablation study on negative sample construction and atomic QA composition strategies, as summarized in Table~\ref{ablation-neg} and Table~\ref{ablation-comp}.

Table~\ref{ablation-neg} investigates the impact of different negative sample strategies. Without any negative samples, the model exhibits limited baseline performance. Introducing cross-image negatives where questions from one image are used as negatives for another significantly degrades the outcome. This suggests that such negatives lack semantic relevance and may introduce misleading noise. In contrast, lexically altered questions as negatives better preserve the semantic structure while introducing controlled perturbations, enabling the model to effectively learn meaningful reasoning boundaries.

Table~\ref{ablation-comp} focuses on atomic QA composition strategies under the lexically altered negative setting. Without composition, the model lacks structural guidance. Fixed template combinations offer a moderate improvement but suffer from deterministic patterns and limited diversity. In contrast, our full method \textit{dual-stage data synthesis approach} achieves the best performance (F1: 84.09\%), highlighting the benefit of structurally diverse and multi-granular supervision. The hierarchical nature of this composition strategy encourages reasoning that better align with human cognition. These findings validate that both semantically meaningful negative samples and structurally diverse reasoning paths are critical to developing interpretable and robust visual reasoning systems.

\subsubsection{Ablation Study of Reward Components}

\begin{table*}[ht]
\vspace{-5pt}
\centering
\renewcommand{\arraystretch}{1.2}
\small
\caption{Ablation study of cognitively coherent verifiable reward (CCVR). Removing any individual reward component leads to a consistent performance drop across all domains.}
\begin{tabular}{lcccccc}
\toprule
\textbf{Method} &
\multicolumn{2}{c}{\textbf{Total}} &
\multicolumn{2}{c}{\textbf{In-domain}} &
\multicolumn{2}{c}{\textbf{Out-of-domain}} \\
& \textbf{ACC(\%)} & \textbf{F1(\%)} & \textbf{ACC(\%)} & \textbf{F1(\%)} & \textbf{ACC(\%)} & \textbf{F1(\%)} \\
\midrule
Baseline & 65.31 & 71.98 & 56.41 & 66.67 & 67.52 & 73.30 \\
\textbf{CCVR (Ours)} & \textbf{77.96} & \textbf{79.60} & \textbf{72.97} & \textbf{75.00} & \textbf{79.19} & \textbf{80.75} \\
\midrule
No format score & 75.81 & 76.92 & 83.78 & 83.33 & 73.83 & 75.47 \\
No answer score & 64.52 & 69.44 & 59.46 & 66.67 & 65.77 & 70.18 \\
No semantic score & 72.58 & 76.92 & 72.97 & 77.27 & 72.48 & 76.84 \\
No edit distance score & 76.34 & 79.63 & 72.97 & 77.27 & 77.18 & 80.23 \\
\bottomrule
\end{tabular}

\label{tab:ccvr_doublecol}
\end{table*}

Results in table \ref{tab:ccvr_doublecol} demonstrate that removing any individual CCVR component—format, answer, semantic, or edit distance—leads to a consistent performance drop across all metrics and domains. The full CCVR model consistently outperforms the baseline across total, in-domain, and out-of-domain splits. Removing any single reward term leads to a performance drop, indicating that the reward components are complementary. Overall, the ablation supports that CCVR’s stepwise reward is important for learning well-structured multi-step reasoning and maintaining robustness.

%% file: section/conclusion.tex
\section{Conclusion and Discussion}
\vspace{-2pt}
Inspired by findings in human cognitive science, we introduce CoTZero, an annotation-free framework that hierarchically parses visual scenes and improves vision–language reasoning via dual-stage data synthesis and cognition-aligned training. CoTZero autonomously generates structured CoT data without human labels and achieves strong performance in both in-domain and out-of-domain settings. Future directions include: (1) exploring more efficient RL optimization algorithms under Cognitively Coherent Verifiable Rewards (CCVR); (2) scaling up synthetic data generation and RL training to further improve VLM reasoning robustness; and (3) extending the hierarchical reasoning paradigm to broader multimodal understanding and reasoning. These advancements could unlock deeper compositional reasoning in VLMs, bridging the gap between pattern recognition and genuine scene comprehension.

%% file: section/Appendix.tex
\begin{figure}
    \centering
    \includegraphics[width=1.0\linewidth]{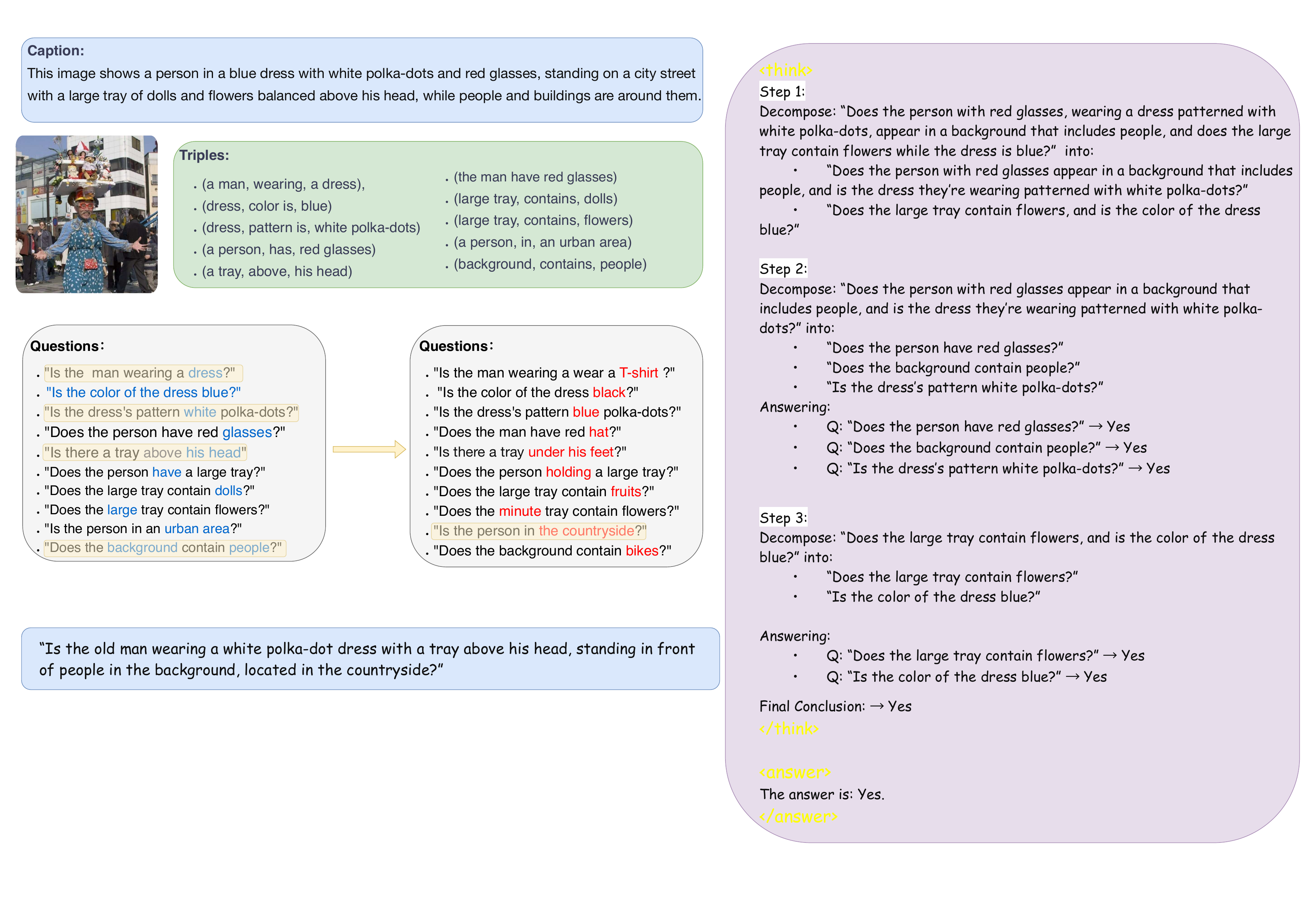}
    \caption{Data generation details.}
    \label{fig:data_gen_details}
\end{figure}
\section{Implementation Details of Data Synthesis} 
\label{app:Data_Generation}

 As illustrated in Fig.~\ref{fig:data_gen_details}, we construct our training dataset through a multi-stage pipeline:
\textbf{Caption Generation.} We use the \textit{Doubao-1.5-Vision-Pro} model to generate concise and factual image descriptions. The system is prompted to use fewer than 150 English words, describe only visible content in the present tense, avoid subjective speculation or technical jargon, and begin with phrases like ``The image'' or ``This is''.

\textbf{Triple Extraction.} We employ \textit{DeepSeek-V3} to extract semantic triples from the generated captions. Each triple is represented as $(\text{Entity}_1, \text{Relation}, \text{Entity}_2)$ and categorized into one of eight relation types: action verbs, state verbs, possession verbs, spatial/location verbs, causality/effect verbs, temporal verbs, quantitative verbs, and perception verbs.

\textbf{Atomic Question Generation and Negative Sampling.} For each triple, we generate an atomic yes/no question. To construct hard negative samples, we replace key tokens in the question such that the answer flips from ``yes'' to ``no''. Each modified question is validated to ensure grammaticality, structural consistency, and successful answer reversal.

\textbf{Compound Question Construction.} Multiple atomic questions are merged into fluent compound questions using \textit{DeepSeek-V3}. A natural language prompt is applied to generate semantically coherent outputs without overusing simple conjunctions such as ``and''.

\section{Dataset Details} 
\label{app:Dataset_Details}

\textbf{Dataset}.We first constructed a training set of 1,000 image-question pairs by sampling a subset of images from FantasyFish, and then expanded it to 4,000 samples by adding images from additional source. These images were annotated using our proposed data generation pipeline, which produces both semantically correct and fine-grained negative questions by modifying key terms based on visual content. For evaluation, we curated a 100-image test set by selecting 20 high-quality samples from each of the following datasets: FantasyFish \cite{laion_art_fantasyfish}, OpenDiffusion \cite{laion2b_en_aesthetic_square_cleaned}, Night2Day \cite{day2night} (night subset), MMInstruction \cite{VLRewardBench}, and LVIS \cite{LAION_LVIS_220}. For each image, we generated negative samples by replacing key terms in the original questions, ensuring that the modified questions remained fluent but semantically incorrect. The test set was structured into multiple difficulty levels, with higher levels involving a greater number of positive distractors. This design allows us to assess the model’s ability to identify subtle semantic inconsistencies under increasingly challenging conditions across both in-domain and out-of-domain samples. 

\subsection{training dataset}
Our training dataset initially consists of 1,000 samples generated using our proposed Annotation-free Hierarchical CoT Generation pipeline, leveraging images from the Fantasy Fish dataset. We then expand the dataset to 4,000 samples, observing consistent improvements in the model’s performance across various aspects.

\begin{figure}
    \centering
    \includegraphics[width=0.5\linewidth]{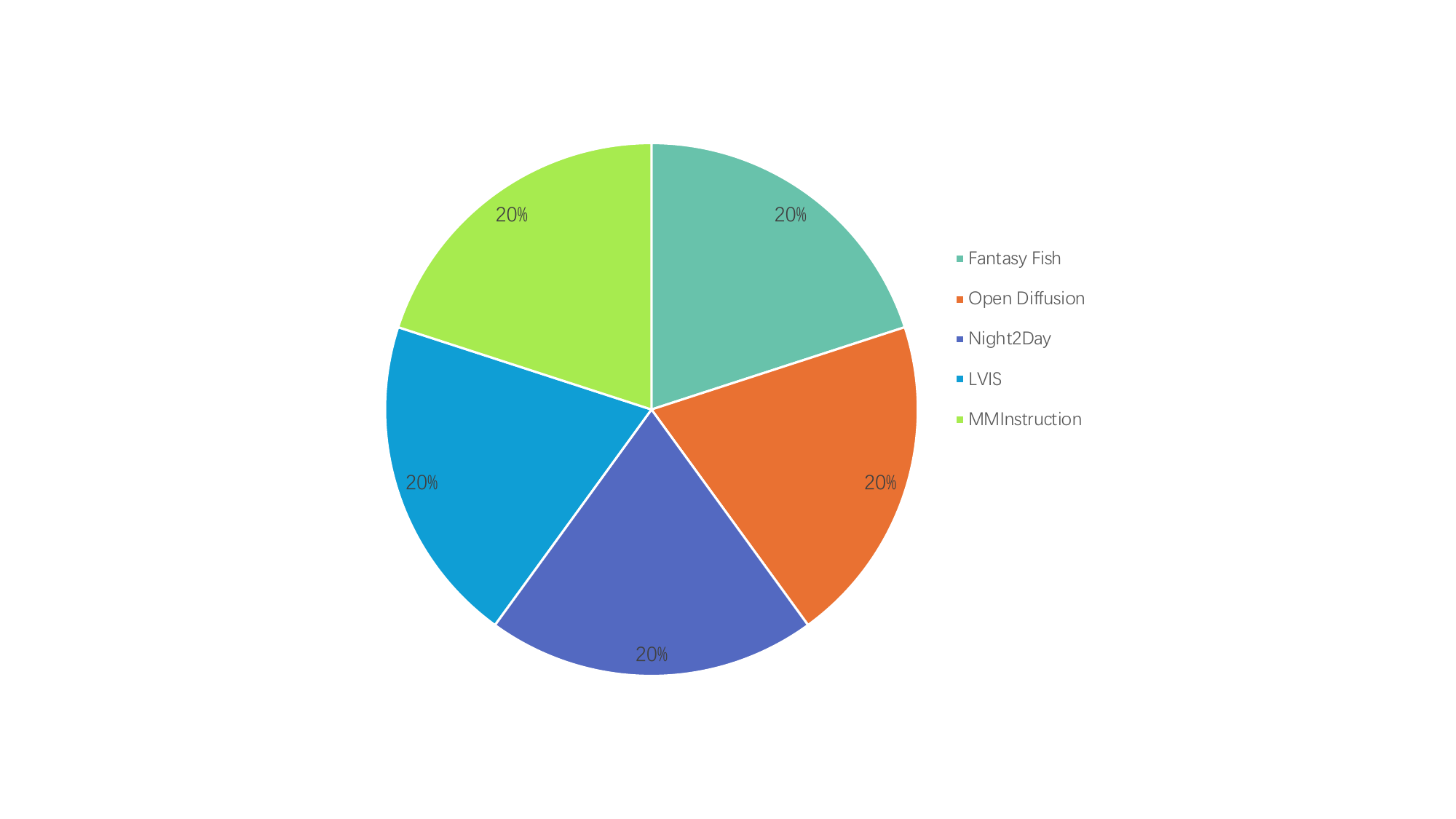}
    \caption{The composition of our test set.}
    \label{data-com}
\end{figure}

\subsection{test dataset}
Figure \ref{data-com} illustrates the composition of our evaluation set, which is uniformly sampled from five sources: FantasyFish \cite{laion_art_fantasyfish}, OpenDiffusion \cite{laion2b_en_aesthetic_square_cleaned}, Night2Day (night subset) \cite{day2night}, MMInstruction \cite{VLRewardBench}, and LVIS \cite{LAION_LVIS_220}, each contributing 20\% of the total. This balanced mixture ensures diversity in object categories, lighting conditions, and instruction-following scenarios, supporting a comprehensive evaluation of our model’s reasoning and perception capabilities.

\clearpage
\section{Additional Quantitative Results} 
\label{app:Additional_Quantitative_Results}
\begin{figure*}[t!]
    \centering
    \includegraphics[width=\linewidth]{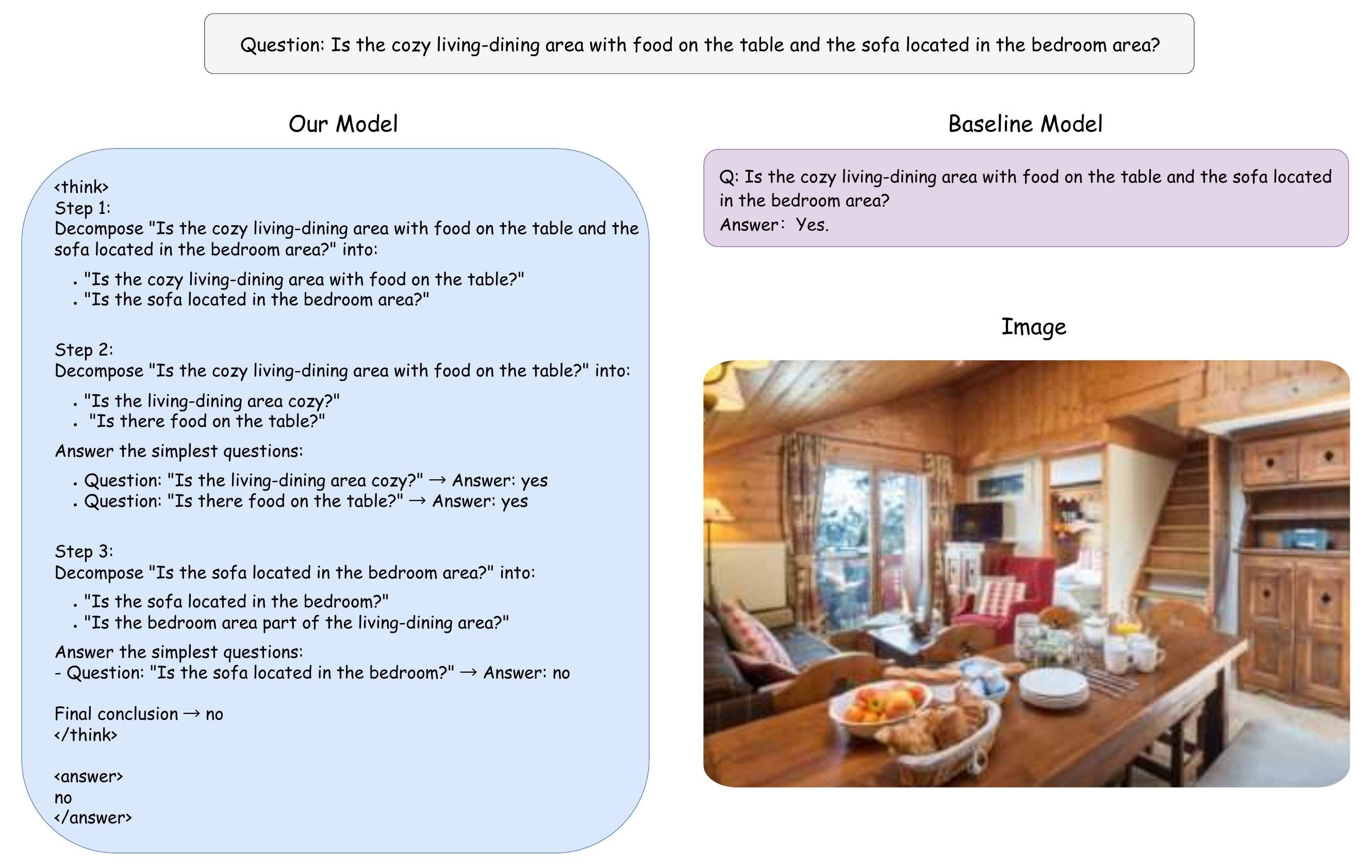}
    \caption{Comparison showing that our model performs step-by-step reasoning to verify details, while the baseline provides a direct answer without decomposition.}
    \label{response_compare}
\end{figure*}

To better understand the differences between our approach and the baseline, we present the comparison in Figure \ref{response_compare}. The baseline model directly outputs a single answer without reasoning, which can lead to errors when the question requires multi-step verification across different visual elements. In contrast, our model decomposes the complex question into sub-questions, systematically verifies each aspect, and then integrates the results to reach a final answer. This structured reasoning enables our model to handle questions involving multiple objects and spatial relationships more reliably, reducing hallucinations and improving consistency in multi-step visual question answering.

\section{Experiment Configuration} 
\label{experiment_config}
\textbf{Experiment configuration}. We first fine-tuned the Qwen2.5-VL-3B model using full-parameter tuning or LoRA with a rank of 64. 
The training is conducted for 3 epochs with a global batch size of 64 on 8 A800 GPUs, using gradient accumulation to simulate large-batch training. 
The learning rate of SFT is set to 2e-5, with a weight decay of 0.1 and cosine learning rate scheduling. A warmup ratio of 0.03 was applied. 
Besides supervised fine-tuning, we further optimize the model via reinforcement learning with the GRPO algorithm. 
The learning rate was set to 1e-6 with a weight decay of 1e-2. An initial KL coefficient of 0.001 was used for regularization. Reinforcement learning was performed for 3 episodes with a batch size of 64.
The model is optimized using DeepSpeed ZeRO3.